\documentclass[sigconf]{acmart}

\usepackage{multirow}
\usepackage{balance}
\AtBeginDocument{%
  }

\copyrightyear{2026}
\acmYear{2026}
\setcopyright{cc}
\setcctype{by}
\acmConference[MM '26]
  {Proceedings of the 34th ACM International Conference on Multimedia}
  {November 10--14, 2026}
  {Rio de Janeiro, Brazil}
\acmBooktitle{Proceedings of the 34th ACM International Conference on
  Multimedia (MM '26), November 10--14, 2026, Rio de Janeiro, Brazil}
\acmISBN{979-8-4007-2213-4/2026/11}
\acmDOI{10.1145/3767308.3836234}

\begin{document}

\title{When Fusion Fails: Corruption-Aware Rebalanced Fusion for Multi-Modal Medical Image Segmentation}

\author{Yuchen Pei}
\email{ycpei@ccnu.edu.cn}
\affiliation{%
  \institution{Central China Normal University}
  \city{Wuhan}
  \country{China}}

\author{Xiaoyu Hu}
\email{xiaoyu\_hu@mails.ccnu.edu.cn}
\affiliation{%
  \institution{Central China Normal University}
  \city{Wuhan}
  \country{China}}

\author{Yixiong Zou}
\email{yixiongz@hust.edu.cn}
\affiliation{%
  \institution{Huazhong University of Science and Technology}
  \city{Wuhan}
  \country{China}}

\author{Dingwen Hu}
\email{20231110467@stu.gzucm.edu.cn}
\affiliation{%
  \institution{Guangzhou University of Chinese Medicine}
  \city{Guangzhou}
  \country{China}}

\author{Hui Chu}
\email{chuhui2019@163.com}
\affiliation{%
  \institution{Guangzhou University of Chinese Medicine}
  \city{Guangzhou}
  \country{China}}

\author{Yutao Ma}
\email{ytma@ccnu.edu.cn}
\affiliation{%
  \institution{Central China Normal University}
  \city{Wuhan}
  \country{China}}

\author{Shijun Qiu}
\correspondingauthor
\email{qiushijun1961@gzucm.edu.cn}
\affiliation{%
  \institution{The First Affiliated Hospital of Guangzhou University of Chinese Medicine}
  \city{Guangzhou}
  \country{China}}

\author{Gang Li}
\correspondingauthor
\email{gang\_li@med.unc.edu}
\affiliation{%
  \institution{University of North Carolina, Chapel Hill}
  \city{Chapel Hill}
  \country{United States}}

\renewcommand{\shortauthors}{Yuchen Pei et al.}

\begin{abstract}
Multi-modal medical image segmentation leverages complementary diagnostic information, yet fusion can underperform single-modality baselines when spatially aligned inputs differ in quality. Here, "corruption" primarily denotes resolution-induced degradation rather than misalignment or complete modality absence, while synthetic noise is evaluated only as an auxiliary setting. We identify a critical optimization-inference inconsistency: degraded modalities can receive weak training updates yet substantially affect predictions, indicating active interference with fusion. We attribute this failure to resampling-induced feature corruption and optimization bias, where noisy features propagate through skip connections and encourage unreliable modality selection. We therefore propose CoReFuse-Med, a Corruption-aware Rebalanced Fusion framework that suppresses corruption during feature transmission and rebalances modality contributions during high-level fusion. Experiments on EPVS, BraTS, and WMH, including multiple Z-axis slice-retention ratios and an auxiliary noise test, demonstrate improved accuracy and robustness under modality-quality discrepancies. Our code is available at https://github.com/lrever/CoReFuse.
\end{abstract}



\maketitle

\section{Introduction}
Multi-modal medical image segmentation (MMIS) has emerged as a cornerstone of modern computer-aided diagnosis, leveraging the synergistic potential of diverse imaging sequences to enhance predictive accuracy and clinical robustness~\cite{zhou2019review,li2023multi,yang2023flexible}. This is because different modalities of images can capture distinct lesion features that do not overlap, thereby jointly improving the definition of the target and achieving better performance compared to single-modal imaging~\cite{zhang2021modality,yao2024drfuse,pereira2016brain,havaei2017brain,soomro2022image}.

Acknowledging the inherent challenges of cross-modal integration, recent literature has extensively addressed explicit forms of heterogeneity. These include spatial misalignment in different modalities~\cite{han2025incomplete,meng2025neighbor}, missing modalities~\cite{wang2023learnable}, and intensity variations across acquisition protocols~\cite{pereira2016brain}. Despite their methodological diversity, these approaches share a common conceptual paradigm: \textbf{fusion failure arises from incomplete or misaligned information}. This perspective has led to a research landscape dominated by alignment, completion, and robust aggregation strategies.

\begin{figure}[!htbp]
    \centering
    \includegraphics[width=\columnwidth]{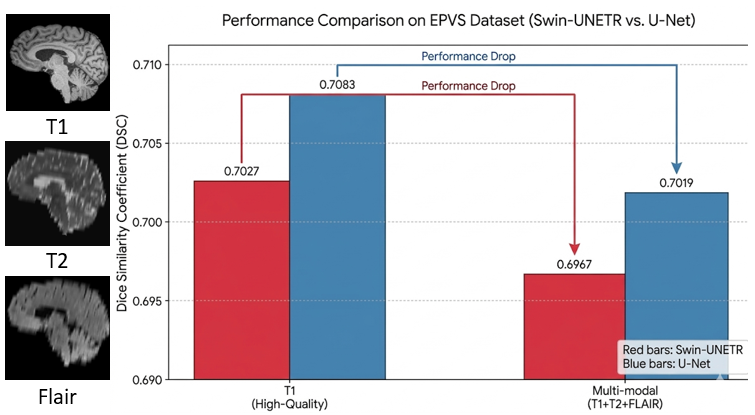}

    \caption{Illustration of multi-modal fusion performance. \textbf{Left:} Visual comparison of high-resolution T1, low-resolution T2, and FLAIR modalities. \textbf{Right:} Quantitative results showing a performance drop in standard architectures when transitioning from T1-only inputs to multi-modal inputs.}
    \Description{High-resolution T1 and lower-resolution T2 and FLAIR brain MRI inputs, together with a quantitative comparison showing that standard multi-modal fusion can underperform a T1-only model.}
    \label{fig:motivation_1}
\end{figure}

However, we identify a distinct and largely unexplored failure mode: \textbf{fusion degradation caused by modality-quality mismatch even when modalities are spatially aligned}. In real-world clinical acquisition, imaging sequences are often captured at different resolutions due to practical constraints such as scan time and patient motion~\cite{muhammad2021comprehensive}. Although standard resampling aligns all modalities into a shared spatial grid, making them appear compatible, we observe a counter-intuitive phenomenon: \textbf{multi-modal fusion can underperform single-modality baselines despite successful alignment}.
As shown in Fig.~\ref{fig:motivation_1}, on the Enlarged Perivascular Spaces (EPVS) MRI dataset with inherent resolution discrepancies (T1: high-resolution; FLAIR/T2: sparse axial slices), multi-modal fusion underperforms the T1 single-modality baseline. This observation shows that additional modalities are not necessarily beneficial when their quality is substantially mismatched, raising a fundamental question: \textit{why does fusion degrade rather than improve performance under aligned quality discrepancies?}

To answer this question, we conduct a systematic analysis from both structural and optimization perspectives. Our empirical findings reveal a critical inconsistency: degraded modalities can exhibit substantial predictive impact despite comparatively weak gradient contributions. This indicates that degraded modalities are neither simply ignored nor reliably exploited; instead, they can substantially bias the decision-making process of the fusion model.

We further identify the root cause of this failure as the interplay between \textbf{feature corruption} and \textbf{optimization bias}. First, resolution discrepancies introduce resampling-induced noise into low-quality modalities, which is subsequently propagated through skip connections and contaminates the fused feature space. Second, under such corrupted inputs, the network tends to adopt a greedy optimization shortcut, over-relying on cleaner modalities while suppressing ambiguous but informative signals from degraded ones. This interplay prevents stable cross-modal reasoning, causing the model to collapse into a noisy and biased modality selection process.

This analysis leads to a key insight: \textit{effective multi-modal fusion under resolution discrepancy conditions requires disentangling two fundamentally coupled challenges, preventing noise propagation and correcting modality-level optimization bias}. This inspires us to design
a corruption-aware fusion network, rather than relying only on the attention-based interactions adopted by current works ~\cite{zhang2021cross,zhou2024shape}. Existing methods primarily focus on feature interaction under the assumption of clean inputs, but overlook modality reliability, allowing noise to propagate and bias to accumulate.

Based on this insight, we propose \textbf{CoReFuse-Med}, a corruption-aware rebalanced fusion framework for imbalanced multi-modal medical image segmentation. Instead of directly enhancing feature interaction, our approach explicitly separates the fusion process into two stages: (1) mitigating feature corruption before fusion to ensure reliable representations, and (2) rebalancing modality contributions during high-level reasoning to avoid biased optimization. This design enables the model to leverage complementary information more effectively under severe resolution discrepancies.

To sum up, our primary contributions are as follows:
\begin{itemize}
\item To the best of our knowledge, we are the first to identify
fusion failure under spatially aligned modality-quality discrepancies, where resolution-induced degradation disrupts cross-modal integration.

\item Gradient and occlusion analyses reveal an optimization--inference inconsistency in which degraded modalities receive weak updates yet strongly affect predictions.

\item We propose \textbf{CoReFuse-Med} that disentangles fusion into two steps, which suppresses shallow feature corruption and rebalances modality contributions during high-level reasoning, enabling reliable multi-modal interaction.

\item Experiments on EPVS, BraTS, and WMH datasets across different modality-corruption settings demonstrate that our method consistently achieves robust performance.
\end{itemize}

\vspace{-10pt}
\section{Related Work}

\textbf{Feature Corruption under Resolution Discrepancy.}
Most multi-modal segmentation networks adopt U-shaped architectures with skip connections to preserve spatial details~\cite{zhang2021modality,zhu2023brain}. While effective for homogeneous inputs, this design implicitly assumes that features transferred from encoder to decoder are reliable. However, in clinical settings, modalities often exhibit significant resolution discrepancies~\cite{muhammad2021comprehensive}. Resampling introduces structured high-frequency noise into low-quality modalities, which is subsequently propagated through skip connections and contaminates the fused process.

Existing multi-modal fusion methods primarily focus on enhancing feature interaction, such as attention-based mechanisms~\cite{zhang2022mmformer,xing2022nestedformer} or resolution-robust operators~\cite{wong2025hnoseg}. These approaches implicitly assume clean and compatible inputs, and therefore tend to amplify, rather than suppress, noise-contaminated signals under cross-resolution conditions. In contrast, our work explicitly attenuates resampling-induced corruption and mitigates its propagation during feature transmission.

\textbf{Optimization Bias in Imbalanced Multi-modal Learning.}
Another line of work addresses modality imbalance, where different modalities contribute unequally due to missing or degraded inputs~\cite{wang2020makes,huang2022modality,fan2023pmr}. It has been shown that deep networks exhibit a greedy optimization tendency~\cite{wu2022characterizing}, over-relying on dominant modalities while suppressing others, a phenomenon often referred to as modality competition.

Recent advances seek to rebalance these dynamics via prototype-based guidance~\cite{fan2023pmr}, self-distillation~\cite{shi2024passion}, or data-level augmentation~\cite{wu2022characterizing,zhou2023adaptive,wei2024enhancing,wang2023learnable}.
However, these methods fundamentally treat modality discrepancy as an issue of data availability or optimization scheduling. They operate under the implicit assumption that the provided modality features are inherently reliable, even if they are weak or incomplete. We argue that this assumption fails in clinical scenarios characterized by resolution discrepancies. In such cases, degraded modalities are not merely under-optimized, and they are intrinsically corrupted by resampling-induced noise.
Hence, they fail to address the coupled effect of feature corruption and optimization bias, where noise-contaminated features further exacerbate modality imbalance during training. Our work instead tackles this problem from a feature-space perspective, jointly suppressing corruption and rebalancing modality contributions during fusion.

\section{Problem Analysis}
\subsection{An Unexpected Finding: Fusion Failure under Aligned Quality Mismatch}
\label{sec:unexpected_finding}

We observe a counter-intuitive phenomenon under spatially aligned modality-quality mismatch: additional modalities do not necessarily improve performance when their quality differs substantially, although complementary modalities are generally expected to benefit segmentation~\cite{zhang2021modality,yang2023flexible}. As shown in Fig.~\ref{fig:motivation_1}, on the IH EPVS dataset with inherent resolution discrepancies (T1: high-resolution; FLAIR/T2: only 10\% axial slices), multi-modal fusion underperforms the single-modality baseline. Specifically, fusing all three modalities yields a DSC of 0.695, which is 1.14\% lower than using T1 alone (0.703). This observation shows that additional modalities are not automatically beneficial under severe quality mismatch, raising a fundamental question: \textit{why does fusion degrade rather than enhance the result?}

Notably, existing multi-modal fusion methods~\cite{zhou2019review,yang2023flexible} do not reveal or explain this phenomenon, as they assume isotropic, high-quality, and well-aligned inputs. This gap motivates our systematic analysis: \textit{why does fusion fail when it should help?}

\subsection{Hidden Noise in Cross-resolution Multi-modal Data}
\label{sec:observation_2}

We hypothesize that the root cause lies in the \textit{resampling-induced noise} arising from cross-resolution discrepancies.
To quantitatively characterize this effect in the latent feature space, we introduce the \textbf{Feature Contrast-to-Noise Ratio} (F-CNR), which quantifies the separability between lesion signals and background noise.

Given an intermediate feature map $F \in \mathbb{R}^{C \times D \times H \times W}$ and a corresponding lesion mask, we compute channel-wise F-CNR by contrasting mean activation in lesion ($\mu_{\text{sig},j}$) and background ($\mu_{\text{bg},j}$), normalized by the background standard deviation:
\begin{equation}
\text{F-CNR}_j = \frac{|\mu_{\text{sig},j} - \mu_{\text{bg},j}|}{\sigma_{\text{bg},j} + \epsilon}
\end{equation}
where $\sigma_{\text{bg},j}$ reflects channel-wise background noise. The final F-CNR is averaged across channels. A higher F-CNR indicates that lesion-related features are more separable from noise-dominated background activations.

\begin{figure}[!htbp]
    \centering
    \includegraphics[width=\columnwidth]{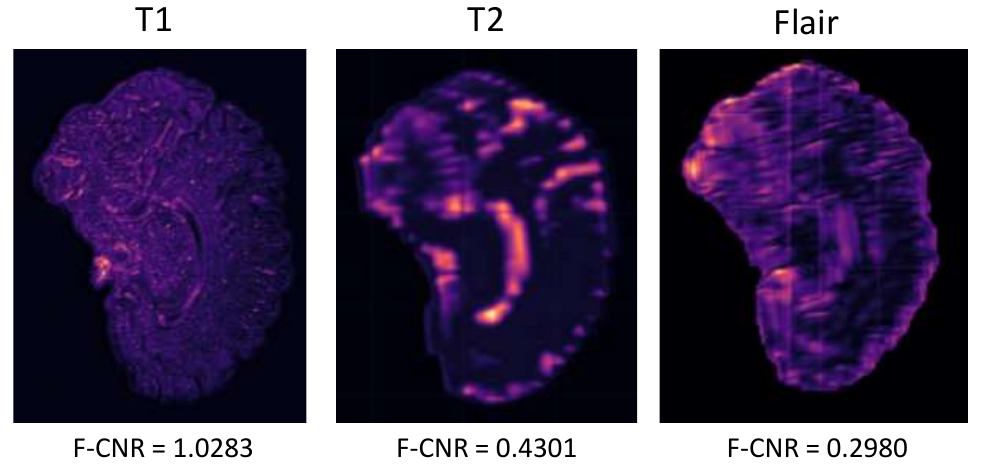}

    \caption{Visualization of shallow features extracted by independent encoders. The T1 modality exhibits high F-CNR and preserves complete structural features, whereas T2 and FLAIR show a significant drop in F-CNR alongside severe structural degradation.}
    \Description{Shallow feature maps from T1, T2, and FLAIR encoders. T1 preserves clearer lesion structure and has a higher feature contrast-to-noise ratio than T2 and FLAIR.}
    \label{fig:noise_analysis}
\end{figure}

As shown in Fig.~\ref{fig:noise_analysis}, the input branches exhibit significant quality disparity.

\textbf{Our Findings.} The F-CNR of T1 is \textbf{3.4$\times$ higher} than that of FLAIR, indicating that resolution discrepancy leads to fundamentally different signal-to-noise characteristics across modalities. Notably, low-resolution modalities (e.g., FLAIR and T2) exhibit elevated background standard deviation $\sigma_{\text{bg}}$, suggesting that their feature responses are strongly affected by resampling artifacts.

Importantly, the resampling-induced corruption analyzed here is \textit{not stochastic}, but structurally induced during resolution alignment. As a result, it systematically corrupts feature representations at early stages and propagates through skip connections into deeper layers. 
\subsection{Fusion Network Takes a Greedy Shortcut}
\label{sec:observation_1}
To understand how noise affects cross-modal fusion, we analyze modality contributions during training via gradient-based attribution. In 3D U-Net, T1 dominates optimization with 77.4\% of gradient contribution, while FLAIR and T2 contribute only 13.1\% and 9.5\%, respectively. However, gradient contribution alone does not reflect the true importance of each modality.

We therefore perform modality-wise occlusion to measure actual predictive impact, as shown in Table~\ref{tab:highlight_modality}. Removing either T1 or FLAIR leads to a similarly large performance drop (DSC $\downarrow$70.2\%), whereas removing T2 results in only minor degradation. This discrepancy indicates that optimization signals are misaligned with actual predictive utility under noisy multi-modal conditions.

Importantly, this phenomenon is not specific to 3D U-Net. In Swin-UNETR, although gradient contributions appear more balanced, occlusion results still reveal substantial differences in modality sensitivity (e.g., DSC $\downarrow$69.6\% for T1 vs. $\downarrow$35.0\% for FLAIR). This suggests that the core issue is not gradient imbalance itself, but the inconsistency between optimization signals and true modality contribution. A similar optimization--inference discrepancy is also observed on BraTS (Table~\ref{tab:highlight_modality}), where gradient contributions do not consistently align with the performance drops caused by modality masking.

As discussed above, degraded modalities are inherently noisy, yet most baseline models rely on U-shaped architectures with skip connections that directly propagate such features. While designed to preserve spatial details, these connections also directly propagate features from encoder to decoder. Under cross-resolution discrepancies, such features are often corrupted by resampling-induced noise, causing degraded modalities to introduce ambiguous and unreliable signals into the fusion process.

Under these conditions, the network exhibits biased optimization dynamics. It tends to favor cleaner signals that are easier to optimize, while noisy features continue to interfere with feature integration. This interplay prevents stable cross-modal reasoning and drives the model toward a greedy shortcut, where it relies on dominant modalities instead of learning robust multi-modal fusion.

\begin{figure*}[!htbp]
    \centering
    \includegraphics[width=\linewidth]{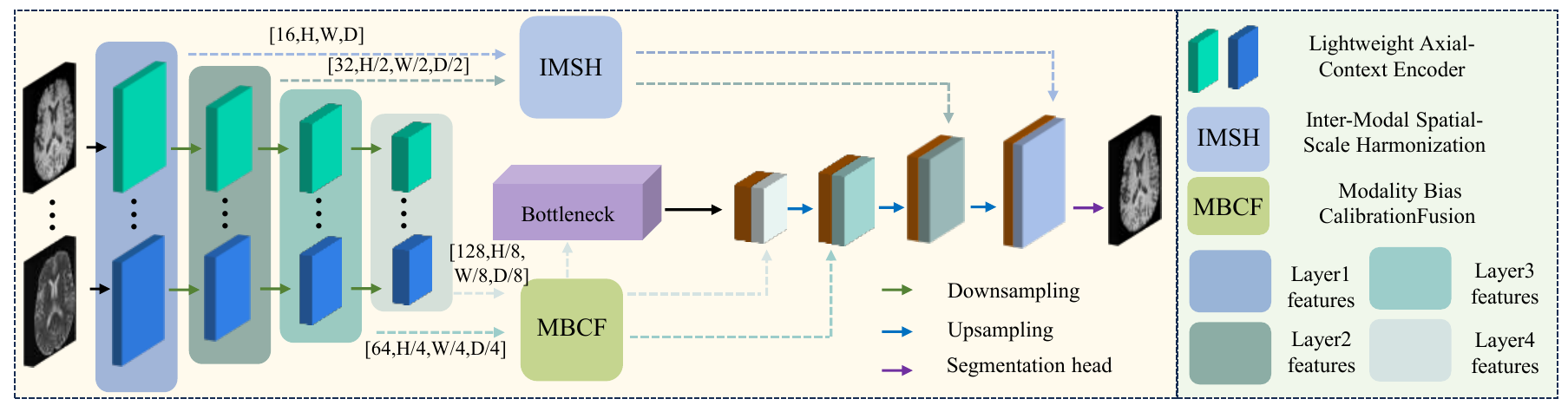}
    \caption{Overview of the CoReFuse-Med Framework. Schematic of the proposed CoReFuse-Med framework, featuring a three-stream symmetric U-shaped architecture with LACE, IMSH and MBCF fusion modules.}
    \Description{A three-stream U-shaped segmentation framework with modality-specific LACE encoders, shallow IMSH fusion, deep MBCF fusion, and a shared decoder.}
    \label{fig:frame_work}
\end{figure*}

\begin{table}[!htbp]
\centering
\caption{Gradient contribution and occlusion-induced DSC drop (\%) on EPVS (M1/M2/M3: T1/T2/FLAIR) and BraTS (T1ce/T2/FLAIR).}
\label{tab:highlight_modality}
\resizebox{\columnwidth}{!}{
\begin{tabular}{llccc|ccc}
\toprule
\multirow{2}{*}{Dataset} & \multirow{2}{*}{Model}
& \multicolumn{3}{c|}{Gradient Contribution}
& \multicolumn{3}{c}{DSC Drop after Masking} \\
\cmidrule(lr){3-5} \cmidrule(lr){6-8}
& & M1 & M2 & M3 & M1 & M2 & M3 \\
\midrule
\multirow{3}{*}{EPVS}
& 3D U-Net & 77.4 & 9.5 & 13.1 & 70.2 & 7.8 & 70.2 \\
& Swin-UNETR & 29.4 & 28.1 & 42.5 & 69.6 & 61.5 & 35.0 \\
& CoReFuse-Med & 91.0 & 2.2 & 6.9 & 73.2 & 0.2 & 0.6 \\
\midrule
\multirow{3}{*}{BraTS}
& 3D U-Net & 49.5 & 19.8 & 30.7 & 55.8 & 10.2 & 40.1 \\
& Swin-UNETR & 40.1 & 28.8 & 31.2 & 49.7 & 13.0 & 60.1 \\
& CoReFuse-Med & 42.6 & 31.7 & 25.7 & 27.6 & 6.2 & 12.3 \\
\bottomrule
\end{tabular}}
\end{table}

The BraTS analysis further shows that a similar gradient--occlusion discrepancy also occurs beyond EPVS. When multiple modalities retain task-relevant evidence, CoReFuse-Med produces a more balanced gradient distribution and substantially smaller occlusion drops than the baselines. Under the extreme EPVS setting, however, T2 and FLAIR retain only sparse through-plane information; their small masking drops suggest limited incremental utility under this severe degradation. The gain in this regime should therefore not be interpreted as reconstructing absent information. Instead, the method prevents contaminated features from harming the dominant T1 pathway while preserving any reliable complementary cues that remain. Together, the two datasets suggest that fusion is limited by both the fusion mechanism and the available task-relevant evidence.

\subsection{Conclusion and Discussion}
The F-CNR results show that resampling changes the signal-to-noise characteristics of degraded modalities at shallow stages, while the gradient--occlusion analysis reveals that this corruption is accompanied by an optimization--inference inconsistency. Skip connections can transmit unreliable features into the decoder, yet the optimization process tends to favor cleaner signals, leaving the final prediction sensitive to modalities that receive limited training updates.

These observations motivate a two-stage solution. IMSH first limits corruption before shallow features enter the shared decoding pathway, and MBCF then calibrates modality contributions in deeper semantic spaces. The goal is not to reconstruct evidence that has been irreversibly removed, but to exploit the remaining complementary information without allowing degraded inputs to destabilize fusion.

\section{Methodology}
\subsection{Framework Overview}

We propose the CoReFuse-Med framework, which introduces a disentangled fusion paradigm for heterogeneous multi-modal medical image segmentation. Unlike conventional strategies that entangle modality-specific artifacts with semantic ambiguities—often leading to sub-optimal optimization—our framework explicitly separates the multi-modal interaction process to address the dual barriers identified in our analysis.

As illustrated in Fig.~\ref{fig:frame_work}, CoReFuse-Med operates hierarchically, built upon the Lightweight Axial-Context Encoder (LACE) for efficient anisotropic feature extraction. In shallow layers, the Inter-Modal Spatial-Scale Harmonization (IMSH) module employs spatial-scale decomposition to isolate shared structural bases from modality-specific resampling artifacts, purifying low-level representations. In deeper semantic spaces, the Modality Bias Calibration Fusion (MBCF) module resolves semantic conflicts via joint channel calibration and symmetric cross-modal attention, effectively preventing the network from over-relying on dominant modalities.

\subsection{Lightweight Axial-Context Encoder}
\label{sec:lace}

\begin{figure}[!htbp]
    \centering
    \includegraphics[width=\linewidth]{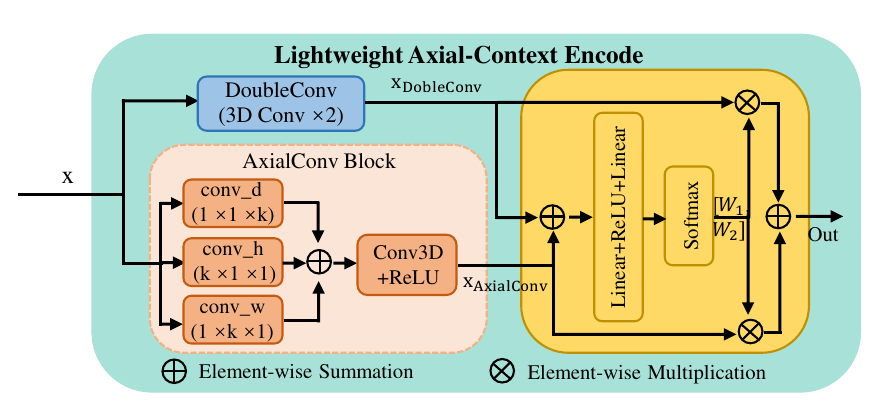}
    \caption{\textbf{Lightweight Axial-Context Encoder.}
Features from DoubleConv and AxialConv branches are integrated via Selective Fusion, where $\oplus$ and $\otimes$ denote element-wise summation and multiplication for generating and applying adaptive attention weights.}
    \Description{The LACE module combines a local convolution branch and factorized axial convolution branches using adaptive selective weights.}
    \label{fig:LACE}
\end{figure}

Clinical MRI volumes are inherently anisotropic, typically exhibiting high in-plane resolution but coarse through-plane spacing, which challenges standard isotropic 3D convolutions. Furthermore, deploying heavy 3D encoders for each modality branch incurs prohibitive parameter growth. To address this, we design the Lightweight Axial-Context Encoder (LACE) as our unimodal backbone (Fig.~\ref{fig:LACE}). LACE employs a dual-branch architecture to concurrently capture local details and direction-aware context—both critical for accurately segmenting elongated and discontinuous lesions—with minimal overhead.

Specifically, the local branch utilizes a DoubleConv block to efficiently extract fine-grained boundaries. In parallel, the axial branch models global context. To avoid the cubic complexity of expanding 3D kernels, it factorizes the volumetric receptive field using parallel 1D strip convolutions along orthogonal axes ($D, H, W$). This explicit decomposition constructs direction-aware context paths, vital for preserving structural continuity across sparsely sampled slices.

To dynamically integrate these complementary local ($\mathbf{X}_{local}$) and axial ($\mathbf{X}_{axial}$) representations, we employ a selective channel-attention mechanism to adaptively balance local detail preservation and global context modeling:
\begin{equation}
\begin{aligned}
    [\mathbf{W}_{local}, \mathbf{W}_{axial}] &= \text{Softmax}(\text{MLP}(\text{GAP}(\mathbf{X}_{local} + \mathbf{X}_{axial}))), \\
    \mathbf{X}_{out} &= \mathbf{W}_{local} \odot \mathbf{X}_{local} + \mathbf{W}_{axial} \odot \mathbf{X}_{axial}.
\end{aligned}
\end{equation}
By structurally decoupling spatial extraction, LACE effectively handles anisotropic geometric variations while maintaining an ultra-lightweight profile compared to conventional 3D encoders.

\subsection{Inter-Modal Spatial-Scale Harmonization}
\label{sec:imsh}

As established in our problem analysis, the spatial resampling required to align different resolution clinical data inevitably introduces aliasing artifacts in low-quality modalities. If standard early fusion (e.g., naive concatenation) is applied, these structured artifacts act as physical noise, significantly degrading the pristine boundary features of the high-resolution modality. To address this, we aim to purify representations before deep semantic interaction.

\begin{figure}[!htbp]
    \centering
    \includegraphics[width=\linewidth]{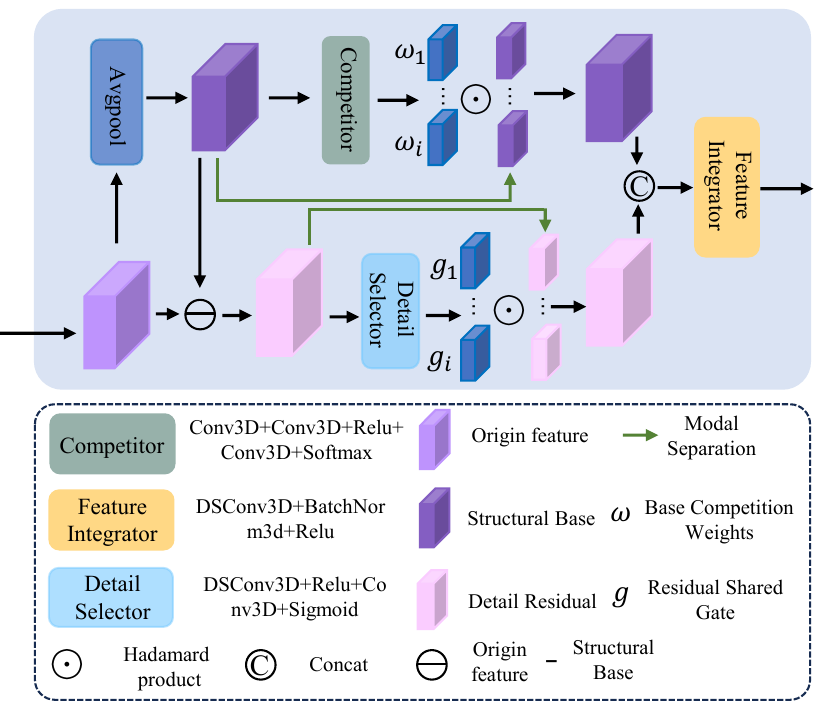}
    \caption{Inter-Modal Spatial-Scale Harmonization. The IMSH module explicitly decouples multi-modal inputs into macroscopic structural bases and detail-sensitive residuals via spatial filtering, suppressing physical noise before deep semantic interaction.}
    \Description{The IMSH diagram separates each modality feature into a smoothed structural base and a detail residual, applies cross-modal competition to the bases and shared gating to the residuals, and reintegrates the filtered components.}
    \label{fig:IMSH}
\end{figure}

We propose the Inter-Modal Spatial-Scale Harmonization (IMSH) module, an efficient fusion mechanism driven by spatial-scale decomposition. Rather than relying on computationally expensive spectral transforms (e.g., FFT-based methods), IMSH achieves feature decoupling via lightweight spatial filtering to isolate shared anatomical layouts from modality-specific artifacts. For multi-modal input features $\{\mathbf{X}_{m}\}_{m \in \{T1, T2, FLAIR\}}$, IMSH operates through three streamlined stages, as illustrated in Fig. \ref{fig:IMSH}.

First, in the spatial-scale decoupling stage, we employ a parameter-free 3D average pooling operator ($\text{AvgPool}_{3\times3\times3}$) as a local smoothing filter to extract the structural base components $\mathbf{X}_{base}$. These components capture the smooth, macroscopic anatomical layout shared across modalities. The corresponding detail-residual components $\mathbf{X}_{residual}$, which contain both fine-grained structural cues and modality-specific resampling noise, are obtained via spatial subtraction:
\begin{equation}
    \mathbf{X}_{base}^m = \text{AvgPool}_{3\times3\times3}(\mathbf{X}_m), \quad \mathbf{X}_{residual}^m = \mathbf{X}_m - \mathbf{X}_{base}^m
\end{equation}

Second, in the base-structure competition stage, we observe that the macroscopic structural components across modalities exhibit strong anatomical consistency. Based on this property, we design a lightweight Competitor module to dynamically aggregate them. To ensure efficiency, the module factorizes standard convolution into depthwise spatial extraction and pointwise channel mixing, followed by a Softmax-based normalization to generate adaptive fusion weights. This allows the model to selectively emphasize the most reliable anatomical layout among modalities while suppressing redundant macroscopic responses.

Third, in the detail-residual shared gating stage, the residual components are highly inconsistent due to modality-specific resampling artifacts. We therefore apply a shared gating function composed of Depthwise Separable 3D Convolutions and a Sigmoid activation. This mechanism adaptively filters out uninformative resampling noise while preserving reliable boundary structures, producing refined residual representations that are robust to modality degradation.

Finally, the scale-specific representations are re-integrated via a single efficient DSConv3d layer to reconstruct the harmonized output representation:
\begin{equation}
    \mathbf{X}_{out} = \mathcal{F}_{final} \left( \left[ \sum_{m} w^m \odot \mathbf{X}_{base}^m, \sum_{m} g^m \odot \mathbf{X}_{residual}^m \right] \right)
\end{equation}

Conventional fusion strategies entangle structural and residual spatial contexts, forcing the network to simultaneously model coarse structural information and fine-grained noise. In contrast, IMSH explicitly separates the structurally consistent macroscopic base from the detail-sensitive residuals via spatial filtering, thereby mitigating cross-modal contamination. This design enables stable cross-modal fusion in shallow layers while maintaining a lightweight computational profile, making it well-suited for practical clinical settings.

\begin{figure}[htbp]
    \centering
    \includegraphics[width=\linewidth]{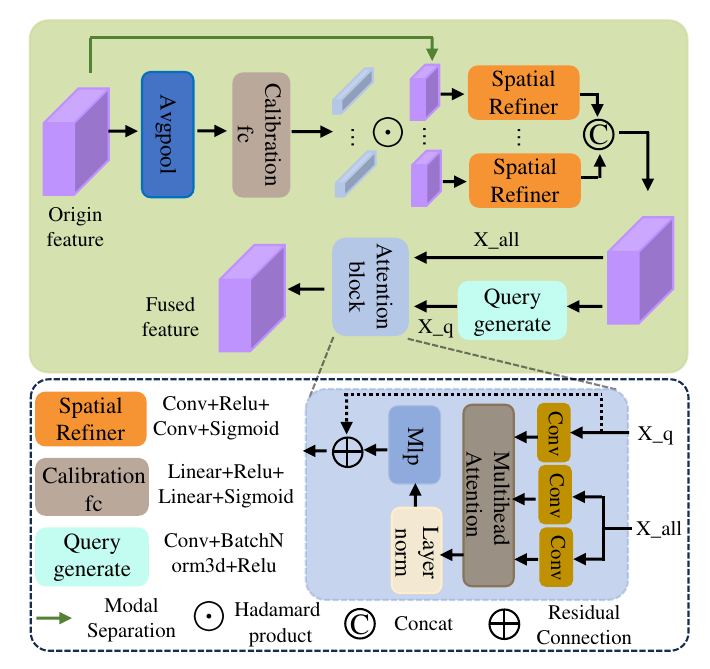}
    \caption{Modality Bias Calibration Fusion. The MBCF module performs hierarchical channel and spatial calibration to resolve deep semantic conflicts between modalities.}
    \Description{The MBCF diagram first calibrates modality channels from joint global descriptors, then applies spatial gates and symmetric cross-attention to obtain a shared fused representation.}
    \label{fig:MBCF}
\end{figure}

\subsection{Modality Bias Calibration Fusion}
\label{sec:mbcf}

As feature representations propagate to deeper layers, they evolve from localized textures to abstract semantics. In this regime, conventional fusion strategies (e.g., concatenation or asymmetric cross-attention) tend to favor modalities with higher signal-to-noise ratio (SNR) or stronger contrast, resulting in biased feature integration.

To address this issue, we propose the Modality Bias Calibration Fusion (MBCF) module, which operates on skip-connected multi-modal features at deep stages (L3--L4), and complements the shallow fusion performed by IMSH at early stages (L1--L2). MBCF performs fusion via joint channel calibration, spatial gating, and symmetric cross-modal attention (Fig.~\ref{fig:MBCF}).

\textbf{Global calibration.}
Global descriptors $\mathbf{v}_m \in \mathbb{R}^C$ are extracted for each modality $m \in \{T1, T2, FLAIR\}$ via global average pooling and concatenated into $\mathbf{v}_{joint} \in \mathbb{R}^{3C}$. A shared MLP produces modality-specific channel weights:
\begin{equation}
[\mathbf{w}_{T1}, \mathbf{w}_{T2}, \mathbf{w}_{FLAIR}] = \sigma(\text{MLP}(\mathbf{v}_{joint}))
\end{equation}

The calibrated features are obtained as $\mathbf{X}_m^{c} = \mathbf{w}_m \odot \mathbf{X}_m$, where $\mathbf{X}_m$ denotes the intermediate skip-connected feature representation of modality $m$ at the current deep stage (L3--L4), and $\mathbf{X}_m^{c}$ denotes the calibrated representation within the same stage. Conditioning the weights on $\mathbf{v}_{joint}$ introduces cross-modal dependency during calibration, which suppresses dominance from modalities with higher SNR or contrast intensity.

\textbf{Spatial gating.}
A $1 \times 1$ convolution followed by a sigmoid function is applied to each calibrated feature map to obtain spatial gates $g_m$, yielding $\mathbf{X}_m^{g} = g_m \odot \mathbf{X}_m^{c}$. The gated features are concatenated and projected into a shared latent space:
\begin{equation}
\mathbf{Q} = \delta(\text{BN}(\text{Conv}_{1\times1}([\mathbf{X}_{T1}^{g}, \mathbf{X}_{T2}^{g}, \mathbf{X}_{FLAIR}^{g}])))
\end{equation}
This projection maps modality-indexed features into a shared latent space while preserving structural semantics at deep semantic levels.

\textbf{Symmetric cross-attention.}
The shared representation $\mathbf{Q}$ is used as query, while concatenated gated features are used as keys and values. To reduce computational cost, keys and values are downsampled via a strided depthwise convolution $\mathcal{F}_{down}$:

\begin{equation}
\mathbf{K}_{joint} = \mathbf{V}_{joint} = \mathcal{F}_{down}([\mathbf{X}_{T1}^{g}, \mathbf{X}_{T2}^{g}, \mathbf{X}_{FLAIR}^{g}])
\end{equation}

The fused representation is computed as:
\begin{equation}
\mathbf{X}_{fused} = \mathbf{Q} + \text{Softmax}\left( \frac{\mathbf{Q}\mathbf{K}_{joint}^T}{\sqrt{d_k}} \right)\mathbf{V}_{joint}
\end{equation}

Eq.~(8) defines a symmetric interaction scheme in which no single modality serves as the query and $\mathbf{Q}$ encodes a shared consensus representation, enabling balanced aggregation across modalities and reducing sensitivity to single-modality dominance. Here, $d_k$ denotes the channel dimension used for scaling in the attention computation.

\begin{table*}[t]
\centering
\footnotesize 
\setlength{\tabcolsep}{4pt} 
\caption{Quantitative comparison on the \textbf{IH EPVS} and \textbf{Synthetic BraTS 2020} datasets. Values denote $\text{mean} \pm \text{std}$. BraTS metrics represent the average across the whole tumor, tumor core, and enhancing tumor. Best and second-best results are \textbf{bolded} and \underline{underlined}, respectively.}
\label{tab:combined_performance}
\begin{tabular}{l|cccc|cccc}
\toprule
\multirow{2}{*}{\textbf{Method}} & \multicolumn{4}{c|}{\textbf{IH EPVS Dataset}} & \multicolumn{4}{c}{\textbf{Synthetic BraTS 2020}} \\
\cmidrule(lr){2-5} \cmidrule(lr){6-9}
 & \textbf{DSC} $\uparrow$ & \textbf{HD95} $\downarrow$ & \textbf{Recall} $\uparrow$ & \textbf{Precision} $\uparrow$ & \textbf{DSC} $\uparrow$ & \textbf{HD95} $\downarrow$ & \textbf{Recall} $\uparrow$ & \textbf{Precision} $\uparrow$ \\
\midrule
\multicolumn{9}{l}{\textit{General-Purpose Backbones}} \\
3D U-Net      & 0.7019 $\pm$ 0.07 & 8.35 $\pm$ 4.66 & 0.7495 $\pm$ 0.06 & 0.6728 $\pm$ 0.11 & 0.7803 $\pm$ 0.17 & 8.43 $\pm$ 9.59 & 0.7639 $\pm$ 0.19 & 0.8647 $\pm$ 0.12 \\
Swin-UNETR    & 0.6967 $\pm$ 0.06 & 8.24 $\pm$ 4.29 & 0.7512 $\pm$ 0.05 & 0.6638 $\pm$ 0.11 & 0.8277 $\pm$ 0.10 & 7.03 $\pm$ 8.37 & \underline{0.8358 $\pm$ 0.12} & 0.8588 $\pm$ 0.10 \\
MedNeXt       & 0.7085 $\pm$ 0.07 & 8.17 $\pm$ 4.09 & \textbf{0.7595 $\pm$ 0.06} & 0.6779 $\pm$ 0.11 & \underline{0.8314 $\pm$ 0.12} & 7.59 $\pm$ 11.29 & 0.8255 $\pm$ 0.13 & \underline{0.8723 $\pm$ 0.08} \\
\midrule
\multicolumn{9}{l}{\textit{Specialized Multi-Modal Networks}} \\
HNoSegXS      & 0.5164 $\pm$ 0.07 & 12.56 $\pm$ 6.50 & 0.4273 $\pm$ 0.09 & 0.6814 $\pm$ 0.08 & 0.8051 $\pm$ 0.14 & 10.30 $\pm$ 12.70 & 0.8004 $\pm$ 0.14 & 0.8498 $\pm$ 0.12 \\
MMFormer      & 0.5444 $\pm$ 0.09 & 29.81 $\pm$ 22.33 & 0.8481 $\pm$ 0.07 & 0.4102 $\pm$ 0.09 & 0.8225 $\pm$ 0.12 & 9.24 $\pm$ 11.23 & \textbf{0.8450 $\pm$ 0.12} & 0.8394 $\pm$ 0.12 \\
\midrule
\textbf{Ours (Base)} & \underline{0.7322 $\pm$ 0.06} & \underline{7.35 $\pm$ 3.90} & 0.7219 $\pm$ 0.11 & \textbf{0.7564 $\pm$ 0.05} & 0.8007 $\pm$ 0.10 & \underline{8.21 $\pm$ 8.11} & 0.7649 $\pm$ 0.12 & 0.8638 $\pm$ 0.07 \\
\textbf{Ours (Med)}  & \textbf{0.7326 $\pm$ 0.06} & \textbf{7.35 $\pm$ 3.89} & \underline{0.7326 $\pm$ 0.08} & \underline{0.7458 $\pm$ 0.05} & \textbf{0.8528 $\pm$ 0.09} & \textbf{4.43 $\pm$ 3.07} & 0.8380 $\pm$ 0.11 & \textbf{0.8978 $\pm$ 0.06} \\
\bottomrule
\end{tabular}
\end{table*}

\begin{table*}[t]
\centering
\footnotesize 
\setlength{\tabcolsep}{6pt} 
\caption{Quantitative comparison on the \textbf{WMH Challenge} dataset. Values denote $\text{mean (interval)}$. Best and second-best results are \textbf{bolded} and \underline{underlined}, respectively.}
\label{tab:wmh_performance}
\begin{tabular}{lccccccc}
\toprule
\textbf{Method} & \textbf{DSC} $\uparrow$ & \textbf{HD95} $\downarrow$ & \textbf{IAVD} $\downarrow$ & \textbf{Recall} $\uparrow$ & \textbf{F1} $\uparrow$ & Para. (M) &GFLOPs \\
\midrule
\multicolumn{8}{l}{\textit{Leaderboard Reference}} \\
sysu\_media (1st) & \textbf{0.80 (0.78-0.82)} & \underline{6.30 (4.75-7.93)} & 0.193 (0.165-0.224) & \textbf{0.84 (0.82-0.86)} & \underline{0.76 (0.73-0.78)} &8.749 & -- \\
\midrule
\multicolumn{8}{l}{\textit{General-Purpose Backbones}} \\
3D U-Net   & 0.77 (0.75-0.79) & 8.24 (6.15-11.05) & 0.216 (0.181-0.256) & 0.75 (0.73-0.77) & 0.68 (0.66-0.71)&22.583 &226.585 \\
Swin-UNETR & 0.77 (0.75-0.79) & 8.24 (6.50-10.27) & 0.211 (0.171-0.259) & 0.80 (0.78-0.82) & 0.67 (0.63-0.70) &61.992 &331.794\\
MedNeXt    & 0.77 (0.75-0.79) & 7.26 (5.84-8.97)  & 0.186 (0.149-0.223) & 0.71 (0.70-0.74) & 0.70 (0.68-0.72)&5.542 & 57.923 \\
\midrule
\multicolumn{8}{l}{\textit{Specialized Multi-Modal Networks}} \\
HNoSegXS   & 0.75 (0.72-0.77) & 11.36 (8.75-14.52) & 0.249 (0.207-0.299) & 0.62 (0.59-0.65) & 0.56 (0.53-0.58) &0.013 &22.636\\
MMFormer   & 0.77 (0.75-0.79) & 8.59 (6.23-11.96)  & 0.216 (0.173-0.258) & 0.79 (0.77-0.81) & 0.66 (0.63-0.69) &9.241 &73.906\\
\midrule
\textbf{Ours (Base)} & 0.78 (0.76-0.80) & 7.37 (5.72-9.26) & \underline{0.182 (0.150-0.217)} & 0.76 (0.74-0.78) & 0.74 (0.72-0.76) & 2.503 &92.162 \\
\textbf{Ours (Med)}  & \textbf{0.80 (0.77-0.81)} & \textbf{6.22 (4.58-8.53)} & \textbf{0.171 (0.139-0.205)} & \underline{0.79 (0.77-0.80)} & \textbf{0.76 (0.74-0.78)} & 2.665 &102.397 \\
\bottomrule
\end{tabular}
\end{table*}

\section{Experiments}
\subsection{Datasets and Implementation Details}

To rigorously evaluate our framework under distinct modality-quality discrepancy scenarios (real-clinical anisotropy, simulated degradation, and information loss), we utilized three datasets.

\textbf{Real Clinical: IH EPVS Dataset.}
This in-house dataset comprises 80 multi-modal MRI scans for Enlarged Perivascular Spaces (EPVS) segmentation. It exhibits inherent clinical anisotropy: FLAIR and T2 sequences contain only 10\% of the axial slices compared to the high-resolution T1 modality.

\textbf{Controlled Simulation: Synthetic BraTS 2020 Dataset.}
Based on BraTS 2020~\cite{mehta2022qu}, we retain T1ce at its original resolution while preserving only 10\% of the original Z-axis slices in FLAIR and T2. The degraded modalities are aligned to the T1ce reference space using FSL-based rigid registration. We further evaluate 20\% and 30\% slice-retention settings, with each model retrained separately under the corresponding condition. Synthetic noise ($\mu=0$, $\sigma=0.25$) is included only as an auxiliary experiment.

\textbf{Information Loss: WMH Challenge Dataset.}
The WMH Challenge dataset~\cite{kuijf2019standardized} features highly discontinuous small lesions with paired T1 and anisotropic FLAIR images. We simulate severe resolution discrepancy by downsampling high-resolution T1 to match the anisotropic FLAIR resolution, evaluating robustness in challenging multi-modal fusion.

\textbf{Implementation Details.}
Models were implemented in PyTorch and trained on an NVIDIA RTX 3090 (24\,GB). EPVS and BraTS used an 8:2 split, while WMH followed the official partitions. Inputs were cropped to $96\times96\times96$, z-score normalized, and optimized with Dice/Focal loss (0.6:0.4).

\subsection{Comparison Methods and Evaluation Metrics}
To rigorously validate the fundamental effectiveness of our proposed mitigating feature corruption and rebalancing modality contributions mechanisms (IMSH and MBCF), we prioritize a transparent comparison against state-of-the-art general-purpose backbones. We select leading paradigms covering pure CNN, Transformer-based, and modernized architectures—specifically 3D U-Net~\cite{cciccek20163d}, Swin-UNETR~\cite{hatamizadeh2021swin}, and MedNeXt~\cite{roy2023mednext}—equipped with standard early fusion. We also compared specialized multi-modal fusion architectures MMFormer~\cite{zhang2022mmformer} and HNoSegXS~\cite{wong2025hnoseg}, a compact resolution-robust segmentation network. Together with the ablation study, this setup helps evaluate whether the observed gains arise from corruption suppression and modality calibration rather than model scale alone.

\textbf{Architectural Versatility.}
To substantiate the generality and architectural agnosticism of our method, we instantiated CoReFuse-Med with two distinct configurations, differentiated by the internal structure of the DoubleConv block: \textbf{Ours (Base)} employs standard convolutions, while \textbf{Ours (Med)} integrates modernized inverted bottleneck blocks inspired by MedNeXt.

\textbf{Evaluation Metrics.}
We report DSC, HD95, recall, and precision on EPVS and BraTS, and additionally report IAVD and F1 following the official WMH benchmark.

\subsection{Quantitative Results and Analysis}
\textbf{Severe Degradation of Specialized Networks.}
Table~\ref{tab:combined_performance} shows that MMFormer and HNoSegXS degrade markedly under severe resolution mismatch. On EPVS, both specialized fusion networks obtain DSC values below 0.55 and substantially larger HD95, while their boundary errors also remain high on BraTS. These results suggest that stronger cross-modal interaction alone does not guarantee robustness when shallow representations contain resampling artifacts. In contrast, \textbf{CoReFuse-Med} achieves HD95 values of 7.35 on EPVS and 4.43 on BraTS, supporting the use of IMSH to filter scale-specific corruption before it enters the decoder through skip connections.

\textbf{Overcoming the Greedy Shortcut.}
On BraTS, MedNeXt exhibits large boundary variance (HD95 $7.59\pm11.29$), indicating unstable predictions across cases under modality degradation. \textbf{CoReFuse-Med} achieves the highest DSC (0.8528) together with a lower HD95 of $4.43\pm3.07$. The simultaneous improvement in overlap and boundary localization suggests that MBCF does more than increase average accuracy: it reduces excessive dependence on the easiest modality and produces more consistent cross-modal interaction when the reliability of the inputs differs.

\textbf{Quantitative Proof of Modality Calibration.}
Table~\ref{tab:highlight_modality} further shows that standard architectures exhibit strong disagreement between gradient contribution and occlusion sensitivity. In EPVS, 3D U-Net loses 70.2\% DSC after masking FLAIR despite its 13.1\% gradient contribution, while Swin-UNETR remains highly sensitive to individual modalities. \textbf{CoReFuse-Med} suppresses severely degraded T2/FLAIR pathways on EPVS, where little recoverable evidence remains, but exhibits more balanced contributions and smaller occlusion drops on BraTS, where multiple modalities retain useful tumor cues. This contrast indicates that calibration is conditioned on task-relevant evidence rather than uniformly suppressing non-dominant modalities.

\textbf{Resolving Semantic Ambiguity for Boundary Fidelity.}
On WMH (Table~\ref{tab:wmh_performance}), \textbf{CoReFuse-Med} achieves HD95 6.22 and IAVD 0.171, compared with 8.59 and 0.216 for MMFormer and 6.30 and 0.193 for the ensemble sysu\_media, demonstrating strong boundary preservation under information loss.

\textbf{Model Complexity and Computational Efficiency.}
\textbf{CoReFuse-Med} uses 2.665M parameters and 102.397 GFLOPs, reducing parameters by 95.7\% and GFLOPs by 69.1\% relative to Swin-UNETR while matching the sysu\_media F1 score (0.76) with a single model.

\begin{table}[t]
\centering
\caption{BraTS performance under varying slice-retention ratios and an auxiliary noise test.}
\label{tab:corruption_robustness}
\renewcommand{\arraystretch}{0.95}
\setlength{\tabcolsep}{3pt}
\resizebox{\columnwidth}{!}{
\begin{tabular}{l|cc|cc|cc}
\toprule
\multirow{2}{*}{Method} & \multicolumn{2}{c|}{Retain 30\%} & \multicolumn{2}{c|}{Retain 20\%} & \multicolumn{2}{c}{Aux. Noise} \\
\cmidrule(lr){2-3} \cmidrule(lr){4-5} \cmidrule(lr){6-7}
& DSC $\uparrow$ & HD95 $\downarrow$ & DSC $\uparrow$ & HD95 $\downarrow$ & DSC $\uparrow$ & HD95 $\downarrow$ \\
\midrule
MedNeXt & 0.850 & 6.69 & 0.839 & 10.16 & 0.811 & 9.75 \\
MMFormer & 0.827 & 8.13 & 0.828 & 9.56 & 0.796 & 10.79 \\
Swin-UNETR & 0.839 & 5.92 & 0.839 & 7.83 & 0.810 & 8.07 \\
\textbf{Ours (Med)} & \textbf{0.864} & \textbf{5.38} & \textbf{0.860} & \textbf{6.99} & \textbf{0.838} & \textbf{6.85} \\
\bottomrule
\end{tabular}}
\end{table}

\textbf{Robustness to Varying Degradation Levels.} Table~\ref{tab:corruption_robustness} extends the extreme 10\% setting to 20\% and 30\% Z-axis slice retention, with each model retrained separately under the corresponding condition. CoReFuse-Med achieves the best DSC and HD95 at both retention ratios, demonstrating consistent effectiveness across different degrees of resolution degradation. The auxiliary noise result provides additional evidence under intensity corruption.

\begin{figure}[htbp]
\centering
\includegraphics[width=\columnwidth]{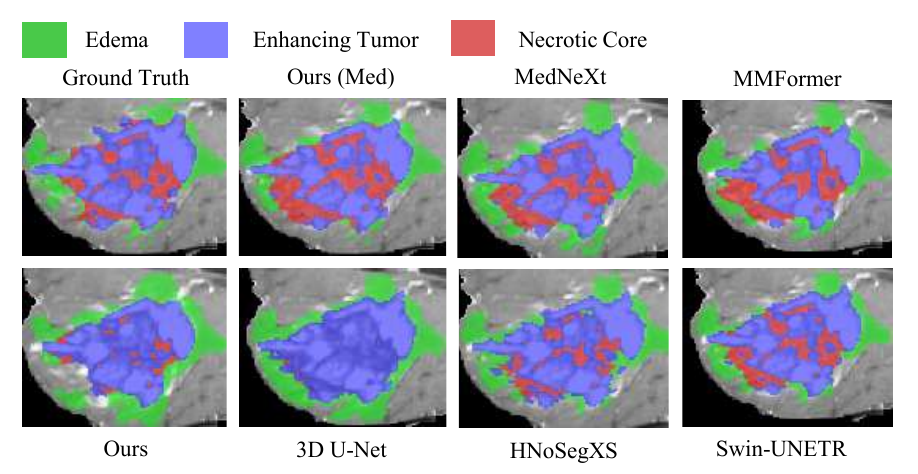}
\caption{Qualitative segmentation results on the Synthetic BraTS 2020 dataset under simulated modality degradation.}
\Description{Qualitative brain tumor segmentation results comparing the proposed method with multiple baselines for edema, enhancing tumor, and necrotic core regions.}
\label{fig:qualitative_segmentation}
\end{figure}

\subsection{Qualitative Visual Analysis}
Figure~\ref{fig:qualitative_segmentation} shows that \textit{Ours (Med)} better preserves irregular edema boundaries and internal tumor structures under degradation. Swin-UNETR and MMFormer tend to oversmooth peripheral regions, while HNoSegXS and 3D U-Net produce fragmented or missed sub-regions. The improvement is most visible around ambiguous transition regions and small internal structures, which are particularly vulnerable to resampling artifacts. Although subtle boundaries remain challenging, the proposed method reduces large fragmented or missing regions and produces more anatomically coherent predictions.

\subsection{Ablation Study}

Table~\ref{tab:ablation_optimized} presents the component analysis on IH EPVS. The baseline achieves relatively high precision but lower recall, indicating conservative predictions dominated by the high-resolution modality. LACE improves DSC and HD95 by strengthening anisotropic contextual modeling. IMSH and MBCF provide larger recall gains by suppressing shallow resampling artifacts and calibrating deep modality interactions, respectively. Combining LACE with either fusion module further reduces boundary errors, while the complete model obtains the best DSC (0.7322) and HD95 (7.35). These results support the complementary roles of shallow spatial harmonization and high-level modality calibration.

\begin{table}[t]
\centering
\footnotesize
\renewcommand{\arraystretch}{0.96}
\setlength{\tabcolsep}{4pt}
\caption{Component ablation on the EPVS dataset.}
\label{tab:ablation_optimized}
\begin{tabular}{l cccc}
\toprule
\textbf{Method} & \textbf{DSC} $\uparrow$ & \textbf{HD95} $\downarrow$ & \textbf{Recall} $\uparrow$ & \textbf{Precision} $\uparrow$ \\
\midrule
Baseline & 0.7146$\pm$0.07 & 8.19$\pm$4.30 & 0.6678$\pm$0.12 & \textbf{0.7908$\pm$0.06} \\
\midrule
\multicolumn{5}{l}{\textit{Single Module Effect}} \\
\quad + LACE & 0.7193$\pm$0.07 & 7.81$\pm$4.50 & 0.6991$\pm$0.12 & 0.7609$\pm$0.06 \\
\quad + IMSH & 0.7219$\pm$0.06 & 7.63$\pm$4.30 & 0.7082$\pm$0.11 & 0.7535$\pm$0.06 \\
\quad + MBCF & 0.7210$\pm$0.06 & 7.88$\pm$4.40 & 0.7090$\pm$0.11 & 0.7499$\pm$0.06 \\
\midrule
\multicolumn{5}{l}{\textit{Combined Effects}} \\
\quad + LACE + IMSH & 0.7287$\pm$0.06 & 7.50$\pm$3.75 & \underline{0.7116$\pm$0.10} & 0.7607$\pm$0.05 \\
\quad + LACE + MBCF & \underline{0.7289$\pm$0.06} & \underline{7.49$\pm$3.90} & 0.7017$\pm$0.11 & \underline{0.7781$\pm$0.06} \\
\midrule
\textbf{Ours (Base)} & \textbf{0.7322$\pm$0.06} & \textbf{7.35$\pm$3.90} & \textbf{0.7219$\pm$0.11} & 0.7564$\pm$0.05 \\
\bottomrule
\end{tabular}
\end{table}

\section{Limitations and Future Work}
Our study primarily addresses spatially aligned resolution degradation rather than complete modality absence, while noise is evaluated only as an auxiliary setting. The in-house EPVS data cannot be released because of clinical privacy constraints, but the implementation and public-dataset protocols will be provided for reproducibility. Future work will extend the framework to dynamically missing modalities and a broader range of acquisition artifacts.

\section{Conclusion}
We present CoReFuse-Med for multi-modal segmentation under spatially aligned modality-quality mismatch, focusing on resolution degradation. By suppressing resampling-induced feature corruption and calibrating modality contributions, CoReFuse-Med improves accuracy and stability across EPVS, BraTS, and WMH, with auxiliary validation under noise. These results highlight the importance of modality reliability when task-relevant information is severely degraded.
\begin{acks}
This work is supported by the Postdoctoral Fellowship Program of China Postdoctoral Science Foundation
(No. GZC20240577, and
2024M751063).
\end{acks}

\bibliographystyle{ACM-Reference-Format}
\balance
\bibliography{sample-base}

\end{document}